\documentclass[11pt,a4paper]{article}
\usepackage[hyperref]{emnlp2020}
\usepackage{times}
\usepackage{latexsym}

\usepackage{microtype}
\usepackage{amsmath}
\usepackage{bm}
\usepackage{color}
\usepackage{multirow}
\usepackage{amsmath}
\usepackage{subfig}
\usepackage{amsfonts}
\usepackage{multirow}
\usepackage{makecell}
\usepackage{arydshln}
\usepackage{graphicx} 
\usepackage{graphics}
\usepackage{bbm}
\usepackage{balance}
\usepackage{algorithm}
\usepackage[noend]{algpseudocode}
\usepackage{booktabs}
\usepackage{textcomp}
\usepackage{threeparttable}
\usepackage{paralist}

\DeclareMathOperator*{\maxpool}{max}
\DeclareMathOperator*{\meanpool}{mean}

\usepackage{CJKutf8}
\usepackage{booktabs}

\usepackage{cleveref}
\crefformat{section}{\S#2#1#3} 

\aclfinalcopy 
\def\aclpaperid{3584} 

\definecolor{jiao}{RGB}{168, 0, 128}

\title{Exploiting Unsupervised Data for Emotion Recognition in Conversations}

\author{Wenxiang Jiao$^\dagger$ ~~~ Michael R. Lyu$^\dagger$ ~~~ Irwin King$^\dagger$  \\
  $^\dagger$~Department of Computer Science and Engineering \\
  The Chinese University of Hong Kong, HKSAR, China \\
  $^\dagger$~{\tt \{wxjiao,lyu,king\}@cse.cuhk.edu.hk} \\}

\date{}

\begin{document}
\maketitle
\begin{abstract}
  Emotion Recognition in Conversations (ERC) aims to predict the emotional state of speakers in conversations, which is essentially a text classification task. Unlike the sentence-level text classification problem, the available supervised data for the ERC task is limited, which potentially prevents the models from playing their maximum effect. In this paper, we propose a novel approach to leverage unsupervised conversation data, which is more accessible.
  Specifically, we propose the Conversation Completion (ConvCom) task, which attempts to select the correct answer from candidate answers to fill a masked utterance in a conversation. Then, we Pre-train a basic COntext-Dependent Encoder (\textsc{Pre-CODE}) on the ConvCom task. Finally, we fine-tune the \textsc{Pre-CODE} on the datasets of ERC. Experimental results demonstrate that pre-training on unsupervised data achieves significant improvement of performance on the ERC datasets, particularly on the minority emotion classes.\footnote{The source code is available at \url{https://github.com/wxjiao/Pre-CODE}}
\end{abstract}

\section{Introduction}
\label{sec:introduction}

\begin{figure}[t]
\centering
    \includegraphics[width=0.9\columnwidth]{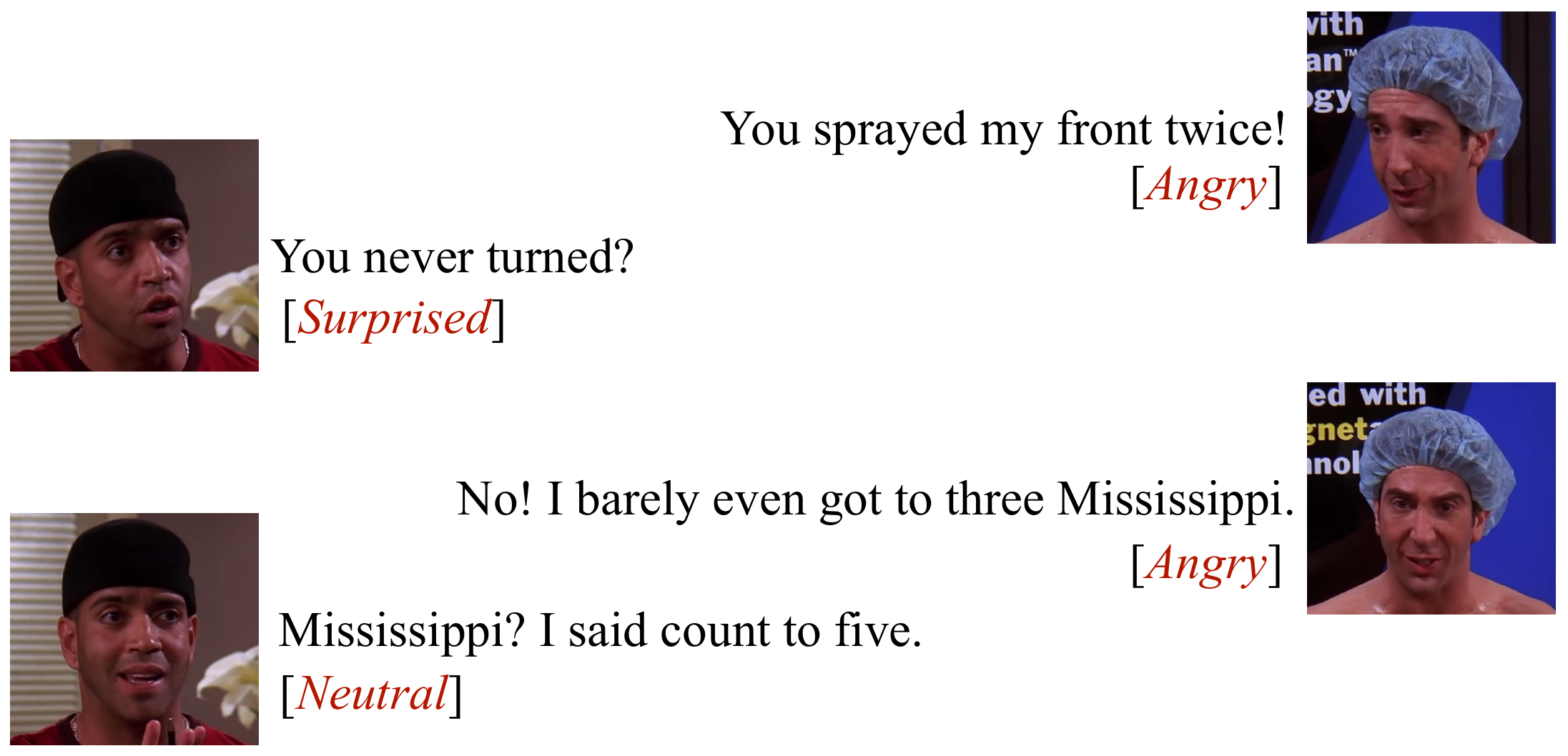}
    \caption{A conversation example with emotion labels.} 
    \label{fig:ERC_example}
\end{figure}

Emotion recognition in conversations (ERC) has garnered attention recently~\cite{poria2019emotion}, due to its potential in developing practical chatting machines~\cite{zhou2018emotional}. Unlike traditional text classification that handles context-free sentences, ERC aims to predict the emotional state of each utterance in a conversation (Figure~\ref{fig:ERC_example}). The inherent hierarchical structure of a conversation, i.e., words-to-utterance and utterances-to-conversation, determines that the ERC task should be better addressed by context-dependent models~\cite{DBLP:conf/acl/PoriaCHMZM17,DBLP:conf/naacl/HazarikaPZCMZ18,jiao2019higru,jiao2020real}. 

Despite the remarkable success, context-dependent models suffer from the data scarcity issue. 
In the ERC task, annotators are required to recognize either obvious or subtle difference between emotions, and tag the instance with a specific emotion label, such that supervised data with human annotations are very costly to collect.
In addition, existing datasets for ERC~\cite{DBLP:journals/lre/BussoBLKMKCLN08,DBLP:conf/acl-socialnlp/HsuK18,DBLP:conf/aaai/ZahiriC18,DBLP:conf/acl/MorencyCPLZ18}
contain inadequate conversations, which prevent the context-dependent models from playing their maximum effect.

In this paper, we aim to tackle the data scarcity issue of ERC by exploiting the unsupervised data. Specifically, we propose the Conversation Completion (ConvCom) task based on unsupervised conversation data, which attempts to select the correct answer from candidate answers to fill a masked utterance in a conversation. Then, on the proposed ConvCom task, we Pre-train a basic COntext-Dependent Encoder (\textsc{Pre-CODE}). The hierarchical structure of the context-dependent encoder makes our work different from those that focus on universal sentence encoders~\cite{DBLP:conf/naacl/PetersNIGCLZ18,radford2018improving,DBLP:conf/naacl/DevlinCLT19}.
Finally, we fine-tune the \textsc{Pre-CODE} on \textit{five} datasets of the ERC task. Experimental results show that the fine-tuned \textsc{Pre-CODE} achieves significant improvement of performance over the baselines, particularly on minority emotion classes, demonstrating the effectiveness of our approach.

Our contributions of this work are as follows:
(1) We propose the conversation completion task for the context-dependent encoder to learn from unsupervised conversation data.
(2) We fine-tune the pre-trained context-dependent encoder on the datasets of ERC and achieve significant improvement of performance over the baselines.

\section{Pre-training Strategy}

\subsection{Approach}
\label{ssec:approach}

\begin{figure}[t]
\centering
    \includegraphics[width=0.9\columnwidth]{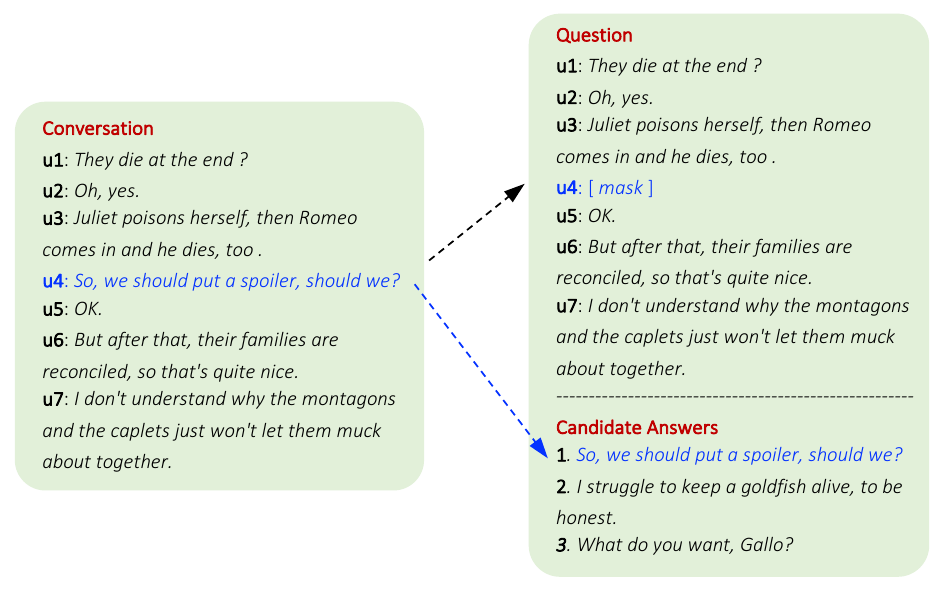}
    \caption{A data example in the ConvCom task.} 
    \label{fig:ConvCom}
\end{figure}

\paragraph{ConvCom Task.}
We exploit the self-supervision signal in conversations to construct our pre-training task.
Formally, given a conversation, $\mathcal{U}=\{ u_1, u_2, \cdots, u_L \}$, we mask a target utterance $u_l$ as $\mathcal{U}\backslash{u_l}=\{ \cdots, u_{l-1}, [mask] , u_{l+1}, \cdots \}$ to create a question, and try to retrieve the correct utterance $u_l$ from the whole training corpus. The choice of filling the mask involves countless possible utterances, making it infeasible to formulate the task into a multi-label classification task with softmax. We instead simplify the task into a response selection task~\cite{DBLP:conf/acl/TongZJM17} using negative sampling~\cite{DBLP:conf/nips/MikolovSCCD13}, which is a variant of noise-contrastive estimation~\cite[NCE,][]{gutmann2010noise}.
To achieve so, we sample $N-1$ noise utterances elsewhere, along with the target utterance, to form a set of $N$ candidate answers. Then the goal is to select the correct answer, i.e., $u_l$, from the candidate answers to fill the mask, conditioned on the context utterances. We term this task ``Conversation Completion", abbreviated as ConvCom.
Figure~\ref{fig:ConvCom} shows an example, where the utterance \texttt{u4} is masked out from the original conversation and the candidate answers include \texttt{u4} and \texttt{two} noise utterances.

\paragraph{Context-Dependent Encoder.}
The context-dependent encoder consists of two parts: an utterance encoder, and a conversation encoder. Each utterance is represented by a sequence of word vectors $\mathbf{X} = \{ \mathbf{x}_1, \mathbf{x}_2, \cdots, \mathbf{x}_T \}$, initialized by the 300-dimensional pre-trained GloVe word vectors\footnote{\url{https://nlp.stanford.edu/projects/glove/}}~\cite{DBLP:conf/emnlp/PenningtonSM14}. 

For the utterance encoder, we adopt a BiGRU to read the word vectors of an utterance, and produce the hidden state 
$\overleftrightarrow{\mathbf{h}}_t = 
[\overrightarrow{\mathbf{h}}_t;
\overleftarrow{\mathbf{h}}_t ] \in \mathbb{R}^{2d_u}$. 
We apply \emph{max-pooling} and \emph{mean-pooling} on the hidden states of all words. The pooling results are summed up, followed by a fully-connected layer, to obtain the embedding of the utterance termed $\mathbf{u}_l$:
\begin{align}
    \mathbf{h}_l &= \maxpool(\{ \overleftrightarrow{\mathbf{h}}_t \}_{t=1}^T) + 
    \meanpool(\{ \overleftrightarrow{\mathbf{h}}_t \}_{t=1}^T), \\
    \mathbf{u}_l &= \tanh(\mathbf{W}_u \cdot \mathbf{h}_l + \mathbf{b}_u), l \in [1, L],
\end{align}
where $T$ denotes the length of the utterance and $L$ is the number of utterances in the conversation. 

For the conversation encoder, since an utterance could express different meanings in different contexts, we adopt another BiGRU to model the utterance sequence of a conversation to capture the relationship between utterances. The produced hidden states are termed $\overrightarrow{\mathbf{H}}_l, \overleftarrow{\mathbf{H}}_l \in \mathbb{R}^{d_c}$.

\begin{figure}[t]
\centering
    \includegraphics[width=1\columnwidth]{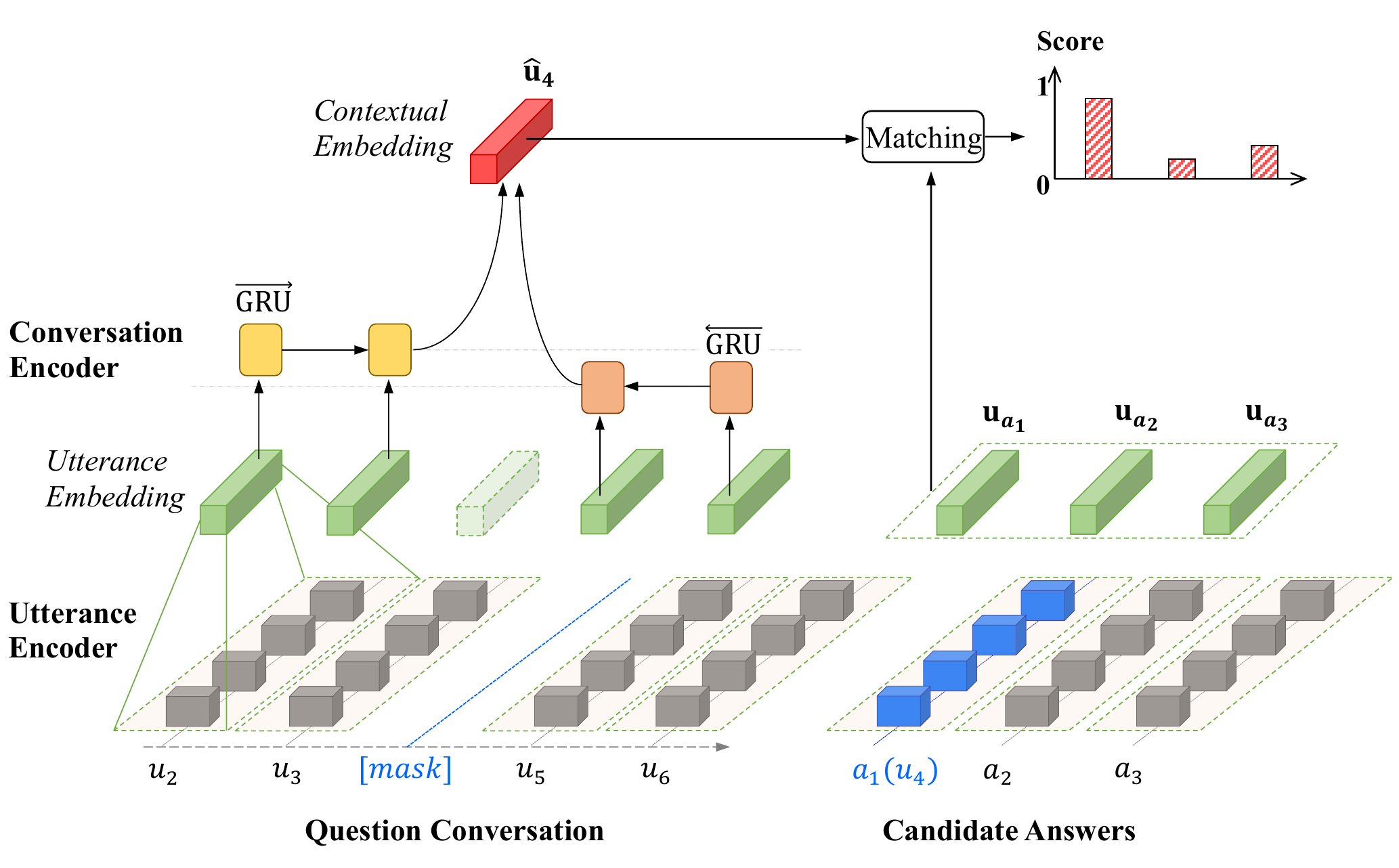}
    \caption{The architecture of the context-dependent encoder with the pre-training objective.}
\label{fig:CODE}
\end{figure}

\paragraph{Pre-training Objective.}
To train the context-dependent encoder on the proposed ConvCom task, we construct a contextual embedding for each masked utterance by combining its context from the history $\overrightarrow{\mathbf{H}}_{l-1}$ and the future $\overleftarrow{\mathbf{H}}_{l+1}$ (see Figure~\ref{fig:CODE}):
\begin{align}
\mathbf{\hat{u}}_l = \tanh(\mathbf{W}_c \cdot [\overrightarrow{\mathbf{H}}_{l-1}; \overleftarrow{\mathbf{H}}_{l+1}] + \mathbf{b}_c) .
\end{align}

Then, the contextual embedding $\mathbf{\hat{u}}_l$ is matched to the candidate answers to find the most suitable one to fill the mask. To compute the matching score, we adopt dot-product with a sigmoid function as:   
\begin{align}
s(\mathbf{\hat{u}}_l, \mathbf{u}_{a_n}) = \sigma(\mathbf{\hat{u}}_l ^\top \mathbf{u}_{a_n}), n \in [1, N],
\end{align}
where $\sigma(x)=\frac{1}{(1+\exp(-x))} \in (0, 1)$ is the sigmoid function, and $\mathbf{u}_{a_n}$ is the embedding of the $n$th candidate answer. The goal is to maximize the score of the target utterance and minimize the score of the noise utterances. Thus the loss function becomes:
{\small
\begin{align}
\mathcal{F} = -\sum_{l}\left[ \log\sigma(\mathbf{\hat{u}}_l ^\top \mathbf{u}_{a_1}) + \sum_{n=2}^{N}\log\sigma(-\mathbf{\hat{u}}_l ^\top \mathbf{u}_{a_n}) \right],
\end{align}
}%
where $a_1$ corresponds to the target utterance, and the summation goes over each utterance of all the conversations in the training set.

\subsection{Experiment}
\label{ssec:experiments_pretrain}

\paragraph{Dataset.}
Our unsupervised conversation data comes from an open-source database OpenSubtitle\footnote{\url{http://opus.nlpl.eu/OpenSubtitles-v2018.php}}~\cite{DBLP:conf/lrec/LisonT16}, which contains a large amount of subtitles of movies and TV shows. 
Specifically, we retrieve the English subtitles throughout the year of 2016, and collect {25,466} \texttt{html} files. After pre-processing, we obtain {58,360}, {3,186}, {3,297} conversations for the training, validation, and test sets, respectively.

\paragraph{Evaluation.}
To evaluate the pre-trained model, we adopt the evaluation metric:
\begin{align}
    \mathbf{R}_{N'}@k = \frac{\sum_{i=1}^{k}y_i}{\sum_{i=1}^{N'}y_i},
\end{align}
which is the recall of the true positives among $k$ best-matched answers from $N'$ available candidates for the given contextual embedding $\mathbf{\hat{u}}_k$~\cite{DBLP:conf/acl/WuLCZDYZL18}. The variate $y_i$ represents the binary label for each candidate, i.e., $1$ for the target one and $0$ for the noise ones. Here, we report $\mathbf{R}_{5}@1$, $\mathbf{R}_{5}@2$, $\mathbf{R}_{11}@1$, and $\mathbf{R}_{11}@2$.

\paragraph{Results.}

\begin{table}[t]
\small
\centering
\begin{threeparttable}
\resizebox{0.98\columnwidth}{!}{
\begin{tabular}{lccccc}
\toprule
\textbf{Model} & $d_u/d_c$ & $\mathbf{R}_{5}@1$ & $\mathbf{R}_{5}@2$ & $\mathbf{R}_{11}@1$ & $\mathbf{R}_{11}@2$ \\
\midrule
\textsc{Small} & 150 & 70.8 & 88.0 & 56.2 & 72.7 \\
\textsc{Mid} & 300 & 73.8 & 89.7 & 60.4 & 76.4 \\
\textsc{Large} & 450 & 77.2 & 91.3 & 64.2 & 79.1 \\
\bottomrule
\end{tabular}
}
\end{threeparttable}
\caption{Test results of CODE on the ConvCom task in three capacities.}
\label{tab:TestPT-CoDE}
\end{table}

For simplicity, we term the context-dependent encoder as CODE. We train CODE on the created dataset in three different capacities, namely, \textsc{Small}, \textsc{Mid}, and \textsc{Large}, corresponding to different hidden sizes of the BiGRUs. See Appendix~\ref{sec:app_pretraining} for the training details.

Table~\ref{tab:TestPT-CoDE} lists the results on the test set. 
For the \textsc{Small} CODE, it is able to select the correct answer for 70.8\% instances with 5 candidate answers and 56.2\% with 11 candidates. The accuracy is considerably higher than random guesses, i.e., 1/5 and 1/11, respectively. By increasing the model capacity to \textsc{Mid} and \textsc{Large}, we further improve the recalls by several points successively. These results demonstrate that CODE is indeed able to capture the structure of conversations and perform well in the proposed ConvCom task.

\section{Fine-tuning Strategy}
\label{sec:fine-tune}

\begin{figure}[t]
\centering
    \includegraphics[width=1.0\columnwidth]{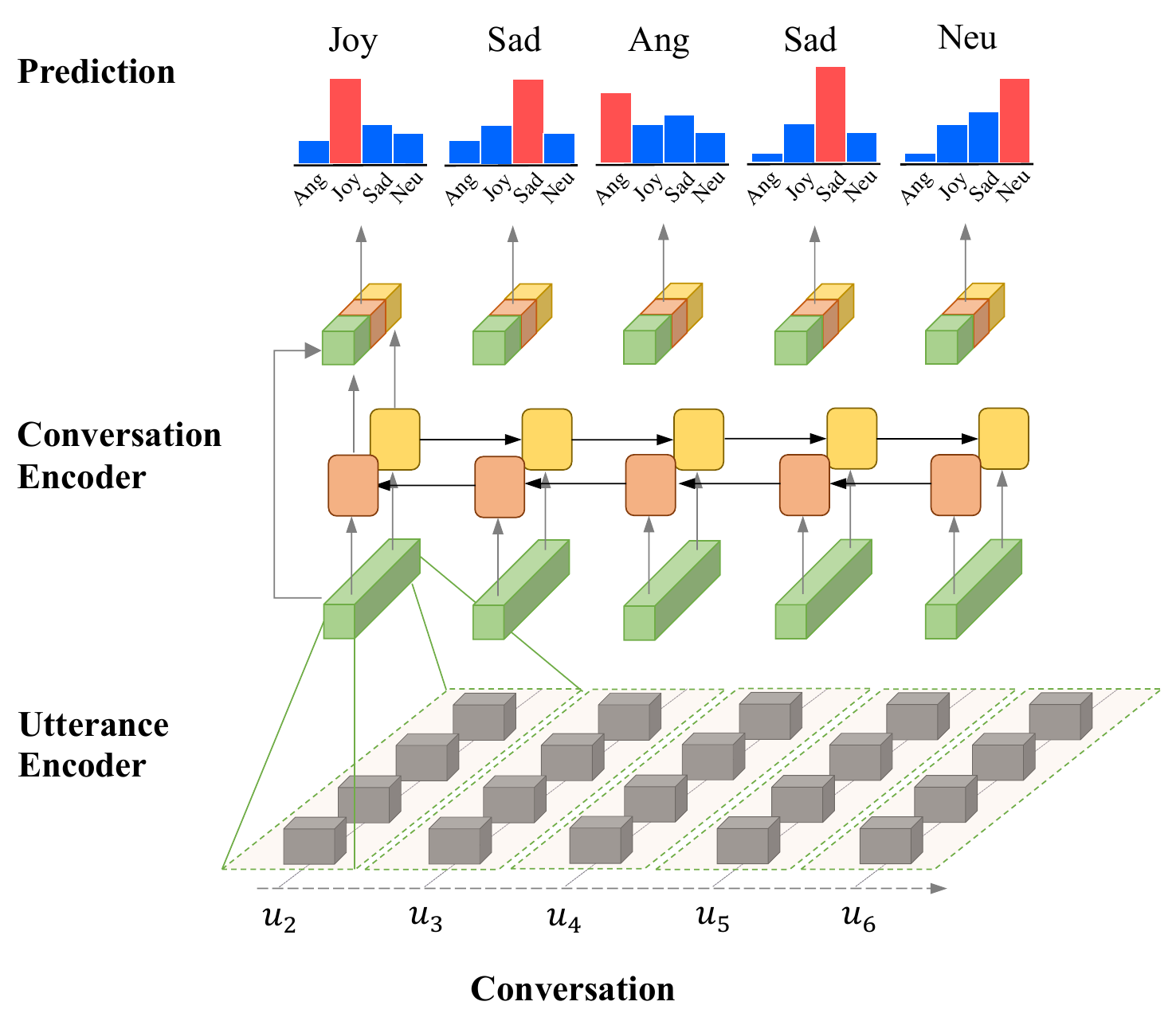}
    \caption{The architecture for the ERC task. Both the utterance encoder and conversation encoder are transferred from the \textsc{Pre-CODE}.}
\label{fig:CODE_ERC}
\end{figure}

\subsection{Experimental Setup}

\paragraph{ERC Architecture.}
To transfer the pre-trained CODE models, termed \textsc{Pre-CODE}, to the ERC task, we only need to add a fully-connected (FC) layer followed by a softmax function to form the new architecture. Figure~\ref{fig:CODE_ERC} shows the resulting architecture, in which we also concatenate the context-independent utterance embeddings to the contextual ones before fed to the FC. 

We adopt a weighted categorical cross-entropy loss function to optimize the model parameters:
{\small
\begin{align}
\mathcal{L} = -\frac{1}{\sum_{i=1}^{N}L_i}\sum_{i=1}^{N} \sum_{j=1}^{L_i} \omega(c_j) \sum_{c=1}^{|\mathcal{C}|} \mathbf{o}_j^c\log_2(\mathbf{\hat{o}}_j^c),
\end{align}
}%
where $|\mathcal{C}|$ is the number of emotion classes, $\mathbf{o}_j$ is the one-hot vector of the true label, and $\mathbf{\hat{o}}_j$ is the softmax output. The weight $\omega(c)$ is inversely proportional to the ratio of class $c$ in the training set with a power rate of 0.5.

\begin{table}[t]
\small
\centering
\resizebox{0.98\columnwidth}{!}{
\begin{threeparttable}
\begin{tabular}{l cc cc cc}
\toprule
\multirow{2}{*}{Model}
& \multicolumn{2}{c}{IEMOCAP}
& \multicolumn{2}{c}{EmoryNLP}
& \multicolumn{2}{c}{MOSEI$^*$}\\
\cline{2-7}
& F1 & WA & F1 & WA & F1 & WA \\
\midrule
bcLSTM\tnote{1} &-- & 73.6 &-- &-- &-- &--\\
CMN\tnote{2} &-- & 74.1 &--&-- &--&-- \\
SCNN\tnote{3} &--&-- & 26.9 & \bf37.9 & -- & --  \\
HiGRU-sf\tnote{4} &--& 82.1 & --&--& --&--\\
\hdashline
bcLSTM & 76.6 & 77.1 & 25.5 & 33.5 & 29.1 & 56.3 \\
bcGRU & 77.6 & 78.2 & 26.1 & 33.1 & 28.7 & 56.4 \\
\hline
\textsc{CODE-Mid} & 78.6 & 79.6 & 26.7 & 34.7 & 29.7 & 56.6 \\
\textsc{Pre-CODE} & \bf81.5 & \bf82.9 & \bf29.1 & 36.1 & \bf31.7 & \bf57.1 \\
\bottomrule
\end{tabular}
\begin{tablenotes}\footnotesize
\item $^1$\newcite{DBLP:conf/acl/PoriaCHMZM17}; $^2$\newcite{DBLP:conf/naacl/HazarikaPZCMZ18};
\item $^3$\newcite{DBLP:conf/aaai/ZahiriC18}; $^4$\newcite{jiao2019higru}.
\end{tablenotes}
\end{threeparttable}
}
\caption{Test results on IEMOCAP, EmoryNLP, and MOSEI$^*$. The implemented bcLSTM performs much better than the original one, possibly because that the original bcLSTM is not trained end-to-end.}
\label{tab:res_IEMOCAP_EmoryNLP_MOSEI}
\end{table}

\begin{table}[t]
\small
\centering
\begin{threeparttable}
\begin{tabular}{l cc cc}
\toprule
\multirow{2}{*}{Model}
& \multicolumn{2}{c}{Friends}
& \multicolumn{2}{c}{EmotionPush}\\
\cline{2-5}
& F1 & WA & F1 & WA \\
\midrule
CNN-DCNN\tnote{1} & -- & 67.0 & -- & 75.7 \\
SA-BiLSTM\tnote{2} & -- & 79.8 & -- & \bf87.7 \\
HiGRU\tnote{3} & -- & 74.4 & -- & 73.8 \\
\hdashline
bcLSTM & 63.1 & 79.9 & 60.3 & 84.8 \\
bcGRU & 62.4 & 77.6 & 60.5 & 84.6 \\
\hline
\textsc{CODE-Mid} & 62.4 & 78.0 & 60.3 & 84.2 \\
\textsc{Pre-CODE} & \bf65.9 & \bf81.3 & \bf62.6 & 84.7 \\
\bottomrule
\end{tabular}
\begin{tablenotes}\footnotesize
\item $^1$\newcite{DBLP:conf/acl-socialnlp/Khosla18}; $^2$\newcite{DBLP:conf/acl-socialnlp/LuoYC18}; 
\item $^3$\newcite{jiao2019higru}.
\end{tablenotes}
\end{threeparttable}
\caption{Test results on Friends and EmotionPush.}
\label{tab:res_Friends_EmotionPush}
\end{table}

\paragraph{Compared Methods.}
We mainly compare our \textsc{Pre-CODE} with bcLSTM~\cite{DBLP:conf/acl/PoriaCHMZM17}, CMN~\cite{DBLP:conf/naacl/HazarikaPZCMZ18}, 
SA-BiLSTM~\cite{DBLP:conf/acl-socialnlp/LuoYC18}, CNN-DCNN~\cite{DBLP:conf/acl-socialnlp/Khosla18},
SCNN~\cite{DBLP:conf/aaai/ZahiriC18},
HiGRU~\cite{jiao2019higru},
and the following: (1) bcLSTM$_{\ddagger}$: bcLSTM re-implemented by us following \citeauthor{jiao2019higru}~\shortcite{jiao2019higru}; (2) bcGRU: A variant of bcLSTM$_{\ddagger}$ implemented with BiGRUs; (3) CODE without pre-training. Unless otherwise stated, CODE and \textsc{Pre-CODE}
are both in the capacity of \textsc{Mid}.

\paragraph{ERC Datasets.}

We conduct experiments on five ERC datasets for the ERC task, namely, 
IEMOCAP~\cite{DBLP:journals/lre/BussoBLKMKCLN08}, Friends~\cite{DBLP:conf/lrec/HsuCKHK18}, 
EmotionPush~\cite{DBLP:conf/lrec/HsuCKHK18}, 
EmoryNLP~\cite{DBLP:conf/aaai/ZahiriC18}, 
and MOSEI~\cite{DBLP:conf/acl/MorencyCPLZ18}. For MOSEI, we pre-process it to adapt to the ERC task and name the pre-processed dataset as MOSEI$^*$ here.
See Appendix~\ref{sec:app_finetuning} for details of the ERC datasets.

\paragraph{Evaluation.}
To evaluate the performance of our models, we report the macro-averaged F1-score~\cite{DBLP:conf/aaai/ZahiriC18}  and the weighted accuracy (WA)~\cite{DBLP:conf/acl-socialnlp/HsuK18} of all emotion classes. The F1-score of each emotion class is also presented for discussion.

\begin{figure}[t]
    \centering
    \subfloat[IEMOCAP]{   \includegraphics[height=0.49\columnwidth]{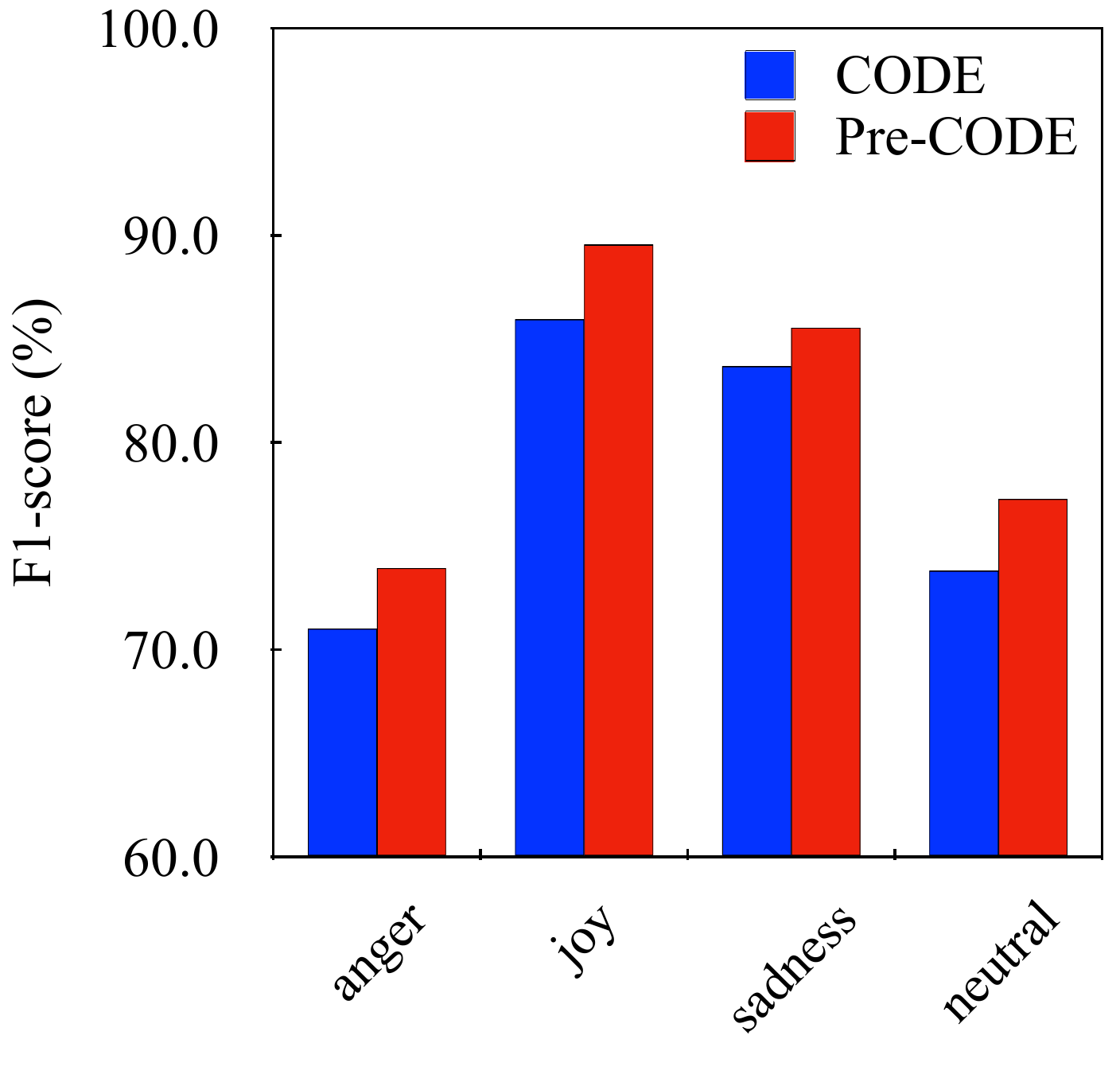}}
    \hfill
    \subfloat[EmoryNLP]{   \includegraphics[height=0.49\columnwidth]{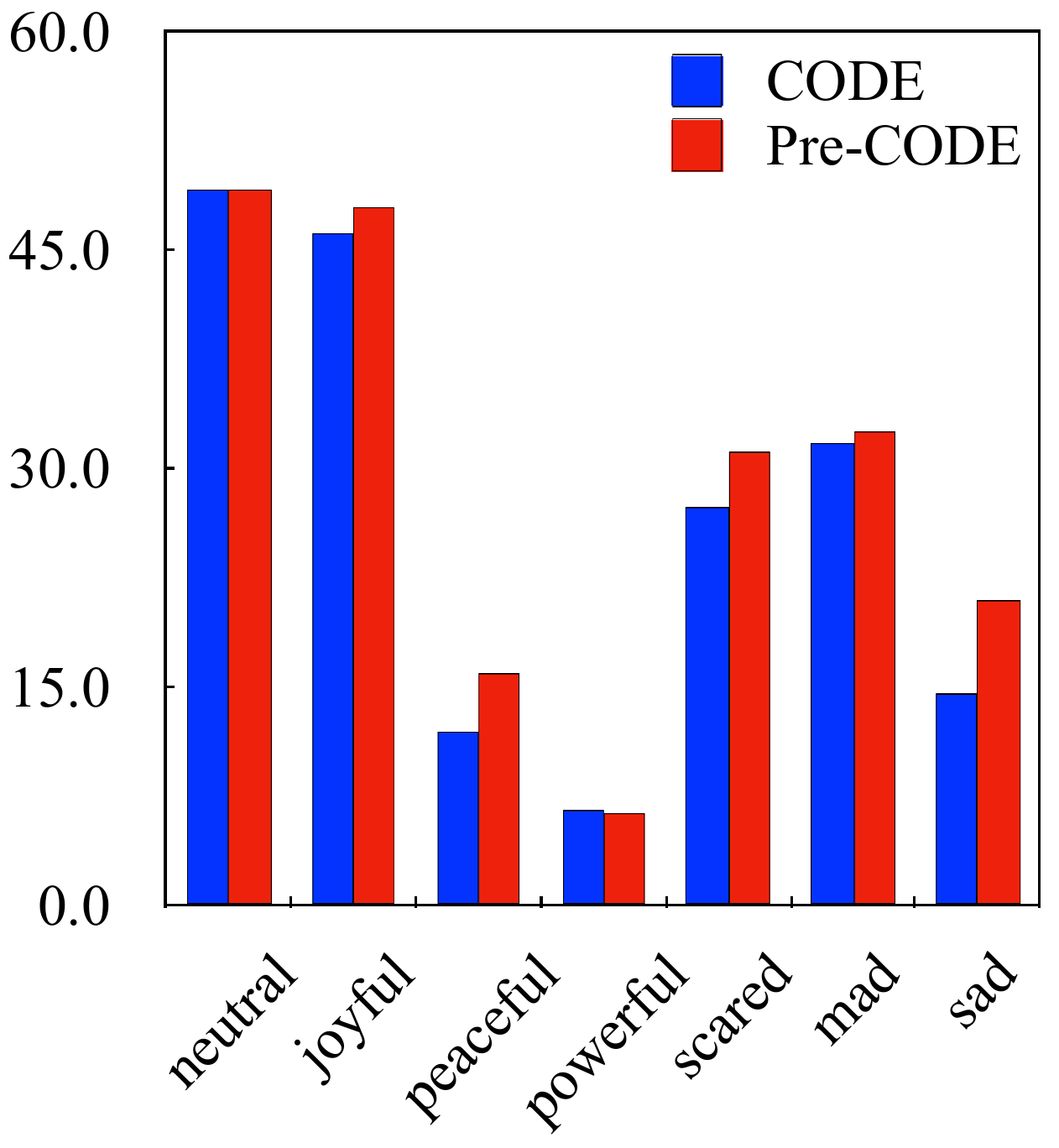}}
    \caption{F1-score of emotion classes on IEMOCAP and EmoryNLP.}
    \label{fig:f1-emotions}
\end{figure}

\paragraph{Results.}
We train the implemented baselines and fine-tune the \textsc{Pre-CODE} on the five datasets. Each result is the average of 5 repeated experiments. See Appendix~\ref{sec:app_finetuning} for training details.

We report the main results in Table~\ref{tab:res_IEMOCAP_EmoryNLP_MOSEI} and Table~\ref{tab:res_Friends_EmotionPush}. As seen, our \textsc{Pre-CODE} outperforms the compared methods on all datasets in terms of F1-score by at least 2.0\% absolute improvement. 
We also conduct significance tests by using two-tailed paired t-tests over the F-1 scores of \textsc{Pre-CODE} and \textsc{CODE-Mid}. P-values are obtained as 0.0107, 0.0038, 0.0011, 0.0003, and 0.0068 for IEMOCAP, EmoryNLP, MOSEI$^*$, Friends, and EmotionPush, respectively. Therefore, the result for IEMOCAP is statistically significant with a significance level of 0.05 whereas the other four datasets obtain a significance level of 0.01.
It demonstrates the effectiveness of transferring the knowledge from unsupervised conversation data to the ERC task.  

To inspect which aspects pre-training helps the most, we present the F1-score of each emotion class on IEMOCAP and EmoryNLP in Figure~\ref{fig:f1-emotions}. As seen, our \textsc{Pre-CODE} particularly improves the performance on minority emotion classes, e.g., \emph{anger} and \emph{sadness} in IEMOCAP, and \emph{peaceful} and \emph{sad} in EmoryNLP. These results demonstrate that pre-training can ameliorate the issue of imbalanced performance on minority classes while maintaining good performance on majority classes.

\subsection{Discussion}

\paragraph{Model Capacity.}
We investigate how the model performance is affected by the number of parameters, as seen in Table~\ref{tab:model_size}. We find that: (1) \textsc{Pre-CODE} consistently outperforms CODE in all cases, suggesting that pre-training is an effective method to boost the model performance of ERC regardless of the model capacity. (2) \textsc{Pre-CODE} shows better performance in the capacities of \textsc{
Small} and \textsc{Mid}, we speculate that the datasets for ERC are so scarce that they are incapable of transferring the pre-trained parameters of the \textsc{Large} \textsc{Pre-CODE} to optimal ones for ERC.

\paragraph{Layer Effect.}
We study how different pre-trained layers affect the model performance, as seen in Table~\ref{tab:layers}. {CODE+Pre-U} denotes that only the parameters of utterance encoder are initialized by \textsc{Pre-CODE}. From CODE to {CODE+Pre-U} and then to \textsc{Pre-CODE}, we conclude that pre-training results in better utterance embeddings and helps the model to capture the utterance-level context more effectively. In addition, {\textsc{Pre-CoDE}+Re-W} represents that we re-train \textsc{Pre-CODE} for 10 more epochs to adjust the originally fixed word embeddings. The results suggest that pre-training word embeddings does not improve the model performance necessarily but may corrupt the learned utterance and conversation encoders.

\begin{table}[t]
\small
\centering
\begin{threeparttable}
\begin{tabular}{l l cc}
\toprule
Model & Capacity & IEMOCAP & Friends \\
\midrule
\multirow{3}{*}{\textsc{CODE}} 
& \textsc{Small} & 76.5 & \bf 62.5 \\
& \textsc{Mid} & \bf 78.6 & 62.4 \\
& \textsc{Large} & 77.6 & 62.1 \\
\cline{1-4}
\multirow{3}{*}{\textsc{Pre-CODE}} 
& \textsc{Small} & 81.2 & 65.2 \\
& \textsc{Mid} & \bf 81.5 & \bf 65.9 \\
& \textsc{Large} & 80.3  & 64.8 \\
\bottomrule
\end{tabular}
\end{threeparttable}
\caption{Ablation study on model capacity.}
\label{tab:model_size}
\end{table}

\begin{table}[t!]
\small
\centering
\begin{threeparttable}
\begin{tabular}{l cc}
\toprule
Layers & IEMOCAP & Friends \\
\midrule
\textsc{Pre-CODE} + Re-W & \bf 81.6 & 64.5 \\
\textsc{Pre-CODE} & 81.5 & \bf 65.9 \\
CODE + Pre-U & 80.1 & 64.8 \\
CODE & 78.6 & 62.4 \\
\bottomrule
\end{tabular}
\end{threeparttable}
\caption{Ablation study on pre-trained layers.}
\label{tab:layers}
\end{table}

\paragraph{Qualitative Study.}
In Table~\ref{table:cases}, we provide two examples for a comparison between CODE and \textsc{Pre-CODE}. The first example is from Friends with consecutive utterances from Joey. It shows that CODE tends to recognize the utterances with exclamation marks ``!" as Angry, while those with periods ``." as Neutral. The problem also appears on \textsc{Pre-CODE} for short utterances, e.g., ``Push!", which contains little and misleading information. This issue might be alleviated by adding other features like audio and video. Still, \textsc{Pre-CODE} performs better than CODE on longer utterances. The other example is from EmotionPush, which are messages with few punctuations. The CODE model predicts almost all utterances as Neutral, which may be because most of the training utterances are Neutral. However, \textsc{Pre-CODE} can identify the minor classes, e.g., Sad, demonstrating that pre-training can alleviate the class imbalance issue.

\begin{table}[t!]
    \centering
    \resizebox{\columnwidth}{!}{
    \begin{tabular}{c p{4cm} ccc}
    \toprule
        \bf Speaker & \quad\quad\quad \bf Utterance & \bf Truth & \bf CODE & \bf \textsc{Pre-CODE} \\
    \midrule
        \it Example 1 & & & & \\
    \hdashline
        Joey & Come on, Lydia, you can do it.				 & Neu & Neu & Neu\\
        Joey & Push!							             & Joy & \colorbox{red!30}{Ang} & \colorbox{red!30}{Ang}\\
        Joey & Push 'em out, push 'em out, harder, harder.   & Joy & \colorbox{red!30}{Neu} & \colorbox{red!30}{Neu}\\
        Joey & Push 'em out, push 'em out, way out!			 & Joy & \colorbox{red!30}{Ang} & Joy\\
        Joey & Let's get that ball and really move, hey, hey, ho, ho.	 & Joy & \colorbox{red!30}{Neu} & Joy\\
        Joey & Let's…  I was just… yeah, right.				 & Joy & \colorbox{red!30}{Neu} & \colorbox{red!30}{Neu}\\
        Joey & Push!							             & Joy & \colorbox{red!30}{Ang} & \colorbox{red!30}{Ang}\\
        Joey & Push!							             & Joy & \colorbox{red!30}{Ang} & \colorbox{red!30}{Ang}\\
    \midrule
        \it Example 2 & & & & \\
    \hdashline
        Sp1 & It's so hard not to cry					     & Sad & \colorbox{red!30}{Ang} & Sad\\
        Sp2 & What happened						             & Neu & Neu & Neu\\
        Sp1 & I lost another 3 set game					     & Sad & \colorbox{red!30}{Neu} & Sad\\
        Sp2 & It’s ok person\_145					         & Neu & Neu & Neu\\
        Sp1 & Why does it hurt so much				         & Sad & \colorbox{red!30}{Neu} & Sad\\
        Sp2 & Everybody loses					             & Neu & Neu & Neu\\
    \bottomrule
    \end{tabular}
    }
    \caption{\label{table:cases} Qualitative comparison between CODE and \textsc{Pre-CODE} by two examples.
    }
\end{table}

\section{Conclusion}
\label{sec:conclusion}

In this work, we propose a novel approach to leverage unsupervised conversation data to benefit the ERC task. 
The proposed conversation completion task is effective for the pre-training of the context-dependent model, which is further fine-tuned to boost the performance of ERC significantly.
Future directions include exploring advanced models (e.g., \textsc{Transformer}) for pre-training, conducting domain matching for the unsupervised data, 
as well as multi-task learning to alleviate the possible catastrophic forgetting issue in transfer learning.

\section*{Acknowledgments}
This work is partially supported by the Research Grants Council of the Hong Kong Special Administrative Region, China (No.~CUHK 14210717, General Research Fund; CUHK 2410021, Research Impact Fund, No. R5034-18).
We thank Xing Wang and the anonymous reviewers for their insightful suggestions on various aspects of this work.

\bibliography{emnlp2020}
\bibliographystyle{acl_natbib}

\appendix

\section{Appendix}

\subsection{Related Work}
\label{sec:app_relatedwork}

Pre-training on unsupervised data has been an active area of research for decades. \citeauthor{DBLP:conf/nips/MikolovSCCD13}~\shortcite{DBLP:conf/nips/MikolovSCCD13} and \citeauthor{DBLP:conf/emnlp/PenningtonSM14}~\shortcite{DBLP:conf/emnlp/PenningtonSM14} lead the heat on learning dense word embeddings over raw text for downstream tasks. \citeauthor{DBLP:conf/conll/MelamudGD16}~\shortcite{DBLP:conf/conll/MelamudGD16}  propose to learn word embeddings in the context with the use of LSTM, which is able to eliminate word-sense ambiguity. More recently, ELMo~\cite{DBLP:conf/naacl/PetersNIGCLZ18} extracts context-sensitive features through a language model and integrates the features into task-specific architectures, achieving state-of-the-art results on several major NLP tasks. Unlike these feature-based approaches, another trend is to pre-train some architecture through a language model objective, and then fine-tune the architecture for supervised downstream tasks~\cite{DBLP:conf/acl/RuderH18,radford2018improving,DBLP:conf/naacl/DevlinCLT19}. With trainable parameters, this kind of approaches are more flexible, attaining better performance than their feature-based counterparts. 

However, the idea of pre-training a context-dependent encoder using unsupervised conversation data for the ERC task has never been explored. On one hand, existing works on ERC focus on modeling the speakers, context, and emotion evolution~\cite{DBLP:conf/acl/PoriaCHMZM17,DBLP:conf/emnlp/HazarikaPMCZ18,DBLP:conf/naacl/HazarikaPZCMZ18,jiao2019higru,jiao2020real}.
No prior work has tried to solve the issue of data scarcity. On the other hand, existing works on transfer learning focus on pre-training universal sentence encoders, e.g., ELMo, GPT, and BERT. But our \textsc{Pre-CODE}, beyond sentence level, is dedicated for sentence sequences from conversations or speeches.
As a result, the pre-training task needs to be customized, for which we propose the ConvCom task. Partially inspired by Word2vec~\cite{DBLP:conf/nips/MikolovSCCD13} and response selection task~\cite{DBLP:conf/acl/TongZJM17}, our ConvCom task differs in that it should model the order of context meanwhile both historical and future context are provided. In contrast, Word2vec neglects the order of context words, and response selection task usually provides only historical context. 

\subsection{Pre-training Strategy}
\label{sec:app_pretraining}

\paragraph{Dataset Creation.}
Our unlabeled conversation data comes from an open-source database named 
OpenSubtitle\footnote{\url{http://opus.nlpl.eu/OpenSubtitles-v2018.php}}~\cite{DBLP:conf/lrec/LisonT16}, which contains a large amount of subtitles of movies and TV shows. 
Specifically, We retrieve the English subtitles throughout the year of 2016, including {25466} \texttt{.html} files. 
We extract the text subtitles from all the \texttt{.html} files and pre-process them as below:
\begin{itemize}
    \item For each episode, we remove the first and the last \textit{ten} utterances in case they are instructions but conversations, especially in TV shows;
    \item We split the conversations in each episode randomly into shorter ones with \textit{five} to \textit{one hundred} utterances, following a uniform distribution;
    \item A short conversation is removed if over half of its utterances contain less than \textit{eight} words each. This is done to force the conversation to capture more information;
    \item All the short conversations are randomly split into a training set, a validation set, and a test set, following the ratio of 90:5:5.
\end{itemize}

Table~\ref{tab:dataset_ConvCom} lists the statistics of resulting sets, where \#Conversation denotes the number of conversations in a set, Avg.~\#Utternace is the average number of utterances in a conversation, and Avg.~\#Word is the average number of tokens in an utterance. Totally, there are over 2 million of utterances in over 60k conversations, which is at least 100 times more than those datasets for ERC (see Table~\ref{tab:dataset_ERC}).

\paragraph{Noise Utterances.}
We randomly sample \textit{ten} noise utterances for each utterance in the training set, validation set, and test set. In each set, a conversation shares the \textit{ten} noise utterances sampled from elsewhere within the set. During training, we can either use the pre-selected noise utterances or sample an arbitrary number of noise utterances dynamically. We use the validation set to choose model parameters, and evaluate the model performance on the test set.

\begin{table}[t!]
\small
\centering
\begin{threeparttable}
\begin{tabular}{l c c c}
\toprule
Set & \#Conversation & Avg.~\#Utterance & Avg.~\#Word \\
\midrule
Train & 58360 & 41.3 & 10.1  \\
Val & 3186 & 41.0 & 10.1 \\
Test & 3297 & 40.8 & 10.1 \\
\bottomrule
\end{tabular}
\end{threeparttable}
\caption{Statistics of the created datasets for the ConvCom task.}
\label{tab:dataset_ConvCom}
\end{table}

\begin{table*}[t]
\centering
\begin{threeparttable}
\begin{tabular}{l ccc ccc}
\toprule
\multirow{2}{*}{Model}
& \multicolumn{3}{c}{\#Conversation}
& \multicolumn{3}{c}{\#Utterance} \\
\cline{2-7}
& Train & Val & Test & Train & Val & Test  \\
\midrule
IEMOCAP & 96 & 24 & 31 & {3,569} & 721 & {1,208} \\
Friends & 720 & 80 & 200 & {10,561} & {1,178} & {2,764} \\
EmotionPush & 720 & 80 & 200 & {10,733} & {1,202} & {2,807} \\
EmoryNLP & 713 & 99 & 85 & {9,934} & {1,344} & {1,328} \\
MOSEI$^*$ & {2,250} & 300 & 676 & {16,331} & {1,871} & {4,662} \\
\bottomrule
\end{tabular}
\end{threeparttable}
\caption{Statistics of the datasets for ERC.}
\label{tab:dataset_ERC}
\end{table*}

\paragraph{Training Details.}
We choose Adam~\cite{DBLP:journals/corr/KingmaB14} as the optimizer with an initial learning rate of $2\times10^{-4}$, which is decayed with a rate of 0.75 once the validation recall $\mathbf{R}_{11}@1$ stops increasing. We use a dropout rate of 0.5 for the utterance encoder and the conversation encoder, respectively. Gradient clipping with a norm of 5 is also applied to avoid gradient explosion.
Each conversation in the training set is regarded as a batch, where each utterance plays the role of target utterance by turns. We randomly sample 10 noise utterances for each conversation during training and validate the model every epoch. The CODE is pre-trained for at most 20 epochs, and early stopping with a patience of 3 is adopted to choose the optimal parameters. Note that, we fix the word embedding layer during pre-training to focus on the utterance encoder and the conversation encoder.

\subsection{Fine-tuning Strategy}
\label{sec:app_finetuning}

\paragraph{ERC Datasets.}
Our \textsc{Pre-CODE} and the implemented baselines are fine-tuned on five ERC datasets, namely, 
IEMOCAP\footnote{\url{https://sail.usc.edu/iemocap/}}~\cite{DBLP:journals/lre/BussoBLKMKCLN08}, Friends\footnote{\url{http://doraemon.iis.sinica.edu.tw/emotionlines}}~\cite{DBLP:conf/lrec/HsuCKHK18}, 
EmotionPush\footnote{\url{http://doraemon.iis.sinica.edu.tw/emotionlines}}~\cite{DBLP:conf/lrec/HsuCKHK18}, 
EmoryNLP\footnote{\url{https://github.com/emorynlp/emotion-detection/}}~\cite{DBLP:conf/aaai/ZahiriC18}, 
and MOSEI\footnote{\url{http://immortal.multicomp.cs.cmu.edu/raw\_datasets/}}~\cite{DBLP:conf/acl/MorencyCPLZ18}. 
For MOSEI, we pre-process it to adapt to the ERC task and name the pre-processed dataset as MOSEI$^*$ here. 
Specifically, we utilize the raw transcripts of MOSEI, where over 14k utterances are not annotated, and others are labeled with one or more emotion labels.
For the unlabeled utterances, we just remove them from the dataset. For the utterance with more than one emotion label, we determine its primary emotion by the majority vote or the highest emotion intensity sum if there are more than one majority votes. For the utterances that obtain zero vote for all emotion classes, we annotate them as \textit{other}.

For the first three datasets, we follow previous work~\cite{DBLP:conf/acl/PoriaCHMZM17,DBLP:conf/lrec/HsuCKHK18} to consider only four emotion classes, i.e., \textit{anger}, \textit{joy}, \textit{sadness}, and \textit{neutral}. We consider all the emotion classes for EmoryNLP as in \cite{DBLP:conf/aaai/ZahiriC18} and six emotion classes (without \textit{neutral}) for MOSEI$^*$.
All the datasets contain the training set, validation set, and test set, except for IEMOCAP. So, we follow \cite{DBLP:conf/acl/PoriaCHMZM17} to use the first four sessions of transcripts as the training set, and the last one as the test set. The validation set is extracted from the randomly-shuffled training set with the ratio of 80:20. We present the statistic details of datasets in Table~\ref{tab:dataset_ERC}.

\paragraph{Training Details.}
We still choose Adam as the optimizer and tune the learning rate for the implemented baselines. Generally, the learning rate of $2\times10^{-4}$ works well for all the datasets except MOSEI$^*$, on which we find $5\times10^{-5}$ works better. For the fine-tuning of \textsc{Pre-CODE}, we use the learning rate of the baselines or its half and report the better results here. We monitor the macro-averaged F1-score of validation set and decay the learning rate once the F1-score stops increasing. The decay rate and patience of early stopping are 0.75 and 6 for all the datasets except IEMOCAP. Since IEMOCAP has much fewer conversations, we change the decay rate and patience of early stopping to 0.95 and 10, respectively.

\end{document}